\documentclass{article}
\usepackage{times}  % Use Times font (recommended for arXiv)
\usepackage{courier}  % Use Courier for monospaced text (code snippets)
\usepackage{graphicx}  % For including figures
\usepackage{hyperref}  % For clickable links
\usepackage{amsmath, amssymb}  % For mathematical symbols
\usepackage{algorithm, algorithmic}  % For pseudocode (if needed)
\usepackage{float}
\usepackage{booktabs}
\usepackage{xcolor}
\usepackage{longtable}
\usepackage{fancyvrb}
\usepackage{subcaption}

\title{Context Discipline and Performance Correlation: Analyzing LLM Performance and Quality Degradation Under Varying Context Lengths}
\author{
Ahilan Ayyachamy Nadar Ponnusamy\\ \texttt{ahilanp@gmail.com}
\and
Karthic Chandran\\ \texttt{karthic.chandran@gmail.com}
\and 
M Maruf Hossain\\ \texttt{maruf.hossain.phd@gmail.com}
}
\date{\today}

\begin{document}

\maketitle

\begin{abstract}
The scaling trend in Large Language Models (LLMs) has prioritized increasing the maximum context window to facilitate complex, long-form reasoning and document analysis. However, managing this expanded context introduces severe computational overhead. This paper investigates the critical trade-off between system performance and model quality when dense transformer architectures--specifically Llama-3.1-70B and Qwen1.5-14B--are exposed to large volumes of irrelevant and distracting context. The research identifies a non-linear performance degradation tied to the growth of the Key-Value (KV) cache. Furthermore, an extended analysis of the Mixture-of-Experts (MoE) architecture reveals unique behavioral anomalies at varying context scales, suggesting that architectural benefits may be masked by infrastructure bottlenecks at high token volumes.

\textbf{Our code and datasets are available on GitHub: }\href{https://github.com/AhilanPonnusamy/LLM-Performance-and-Quality-Degradation-with-Context-Length}{GitHub Repository}.
\end{abstract}

\section{Introduction}
The practical deployment of Large Language Models (LLMs) hinges on balancing model capability with serving efficiency. While models now routinely support context windows exceeding 100,000 words, system resources, particularly GPU, VRAM, and memory bandwidth, are directly consumed by the Key-Value (KV) cache associated with these lengthy inputs. This inherent cost structure makes the length of the input context a critical performance multiplier in production environments.

As the industry moves toward ``Long-Context'' applications, a fundamental question arises: is the impact of long context purely a matter of computational latency, or does the presence of ``noise'' fundamentally compromise the model’s reasoning capabilities? This study focuses on two key inquiries:
\begin{enumerate}
\item \textbf{Performance Impact:} The severity of inference time scaling when processing distracting context up to 15,000 words is compounded by the fact that tokens are not synonymous with words, but are rather sub-word fragments that can vary in length. Because different models use distinct vocabularies and tokenization algorithms, a 15,000-word limit can represent significantly more or less actual text depending on whether a model’s tokenizer is optimized for efficiency or for specific languages.
\item \textbf{Quality Degradation:} Whether reasoning accuracy and retrieval capabilities are compromised by the presence of long context and the positioning of the information in the context, even when the required information is present within the input.
\end{enumerate}

We address these questions by conducting a rigorous benchmarking analysis across dense transformer architecture LLMs: \textbf{Llama-3.1-70B-Instruct} and \textbf{Qwen1.5-14B-Chat}, and then attempted to replicate the experiment using a Mixture-of-Experts (MoE) architecture model \textbf{Mixtral-8x7B-Instruct-v0.1} with approximately 13B active parameters.

\section{Related Work}
Optimizing LLM inference requires addressing the quadratic scaling of attention and the memory bottlenecks inherent to long-sequence processing. Several distinct research directions inform our work:

\begin{itemize}
\item \textbf{Attention Efficiency and IO-Awareness:} A primary challenge in scaling context is the ``memory wall''. Dao \textit{et al.}~\cite{dao2022flashattentionfastmemoryefficientexact} demonstrated that standard attention is often bottlenecked not by compute, but by memory reads/writes between GPU HBM and SRAM. Their FlashAttention algorithm addresses this by tiling, thereby reducing I/O complexity. While our study focuses on the system-level impacts of context distraction, the latency floors we observe are fundamentally governed by these IO-aware principles.
\item \textbf{KV Cache Management and Scaling:} Pope \textit{et al.}~\cite{pope2022efficientlyscalingtransformerinference} investigated the engineering tradeoffs for 500B+ parameter models, highlighting that multiquery attention (MQA) enables scaling to 32x larger context lengths by reducing KV cache memory requirements. Furthermore, Li \textit{et al.}~\cite{li2025surveylargelanguagemodel} provide a comprehensive taxonomy of KV cache management, from token-level quantization to system-level scheduling. Our research builds on these systems-level insights by empirically measuring how these memory-bound architectures respond when the KV cache is saturated with ``noise''.
\item \textbf{Positional Bias and Utilization:} Despite the ability to process long sequences, models do not always utilize the context uniformly. Liu \textit{et al.}~\cite{liu2023lostmiddlelanguagemodels} identified the ``Lost in the Middle'' phenomenon, where performance peaks when relevant information is at the extreme start or end of the input but degrades significantly in the center. Our ``Needle in a Haystack'' methodology utilizes these findings to benchmark whether modern dense transformer models have improved their robustness to such positional biases.
\item \textbf{Sparse Architectures and Reliability:} Sparse activation models, such as the Switch Transformer proposed by Fedus \textit{et al.}~\cite{fedus2022switchtransformersscalingtrillion}, aim to increase parameter counts while keeping compute costs constant. However, as our study reveals in the MoE extension, these models still face significant KV cache overheads at extreme context lengths. Additionally, the reliability of these models under misinformed or noisy prompts, as explored by Aremu \textit{et al.}~\cite{aremu2024reliabilitylargelanguagemodels}, is a critical factor in deployment. We extend this by showing that ``noise'' is not just a qualitative risk but a quantifiable performance penalty.
\end{itemize}

\section{Methodology and Experimental Focus}
The experimental framework was designed to quantify the relationship between context size and quality across three context regimes: 4,096, 10,000, and 15,000 words.

\subsection{Primary Testing Methodologies}
\begin{itemize}
\item \textbf{Context Distraction Test:} Models were tasked with answering 200 diverse general knowledge questions. To simulate real-world data noise, each question was preceded by a large, irrelevant context block, forcing the model to distinguish relevant signals from distracting information.
\item \textbf{Needle in a Haystack Test (NIAH):} This measured the ability to retrieve a single, specific fact hidden within the ``haystack'' of distracting context. The position of the ``needle'' was varied to ensure results were not skewed by the positional biases identified by Liu \textit{et al.}~\cite{liu2023lostmiddlelanguagemodels}.
\end{itemize}

\subsection{System Architecture and Environment}
Testing was performed on an \textbf{AWS p4d.24xlarge} instance featuring the following configuration. This is a high-performance computing instance designed for training and deploying large-scale machine learning models.

\subsubsection*{Key Configuration:}
\begin{itemize}
    \item \textbf{GPUs:} 8 NVIDIA A100 Tensor Core GPUs.
    \item \textbf{GPU Memory (VRAM):} 40 GB per GPU (Total 320 GB VRAM).
    \item \textbf{System Memory (RAM):} 1.1 TB (1,152 GB).
    \item \textbf{vCPUs:} 96 vCPUs (Intel Xeon Scalable processors).
    \item \textbf{Network Bandwidth:} 400 Gbps.
\end{itemize}

The models were served using the \textbf{vLLM inference engine}, which utilizes PagedAttention to manage non-contiguous memory spaces for the KV cache~\cite{li2025surveylargelanguagemodel}. To ensure raw performance was measured, prefix caching was disabled, and a tensor-parallel size of 8 was used to accommodate the 70B model's large parameter count.

An example configuration would look like the following:

\subsubsection*{Llama-3.1-70B-Instruct}
\begin{Verbatim}[fontsize=\small]
python3 -m vllm.entrypoints.openai.api_server \
    --model meta-llama/Llama-3.1-70B-Instruct \
    --max-model-len 128000 \
    --host 0.0.0.0 \
    --port 8000 \
    --no-enable-prefix-caching \
    --max-num-batched-tokens 128000 \
    --tensor-parallel-size 8 \
    --kv-cache-memory-bytes 14000000000
\end{Verbatim}

The details and importance of these parameters are shown in Tab.~\ref{tab:vllm-parameters}:

\begin{table}[h!t]
\centering
\small
\begin{tabular}{p{0.18\textwidth}p{0.35\textwidth}p{0.37\textwidth}}
\hline
\textbf{Parameter} & \textbf{Explanation} & \textbf{Necessity for Testing} \\
\hline
\textbf{--model} & 
\textbf{Model Identifier.} Specifies the path or Hugging Face ID of the model to load 
(meta-llama/Llama-3.1-70B-Instruct in this case). & 
\textbf{Necessary.} You cannot start the server without defining the model. \\
\hline
\textbf{--max-model-len} & 
\textbf{Max Context Length.} Defines the maximum number of tokens (input + output) the 
model will accept. This directly dictates the memory allocated for the KV cache 
per request. & 
\textbf{Necessary.} Setting this correctly is fundamental to testing the context lengths 
specified in the Canvas (4k, 10k, 15k, etc.). \\
\hline
\textbf{--host / --port} & 
\textbf{Network Configuration.} Defines the network interface (0.0.0.0 or 127.0.0.1) and 
the port (8000) the API listens on. & 
\textbf{Necessary.} Required to make the server accessible to your client script. \\
\hline
\textbf{--no-enable-prefix-caching} & 
\textbf{Prefix Caching Control.} Disables the feature where vLLM caches shared prompt 
prefixes across requests. & 
\textbf{Recommended.} Disabling this simplifies the memory management during testing and 
ensures each thread's context is treated as unique, providing a cleaner test of 
raw concurrent load. \\
\hline
\textbf{--max-num-batched-tokens} & 
\textbf{Maximum Batch Size.} The maximum number of tokens (including both context and 
generated tokens) that can be processed simultaneously in a single GPU kernel run. & 
\textbf{Necessary.} Setting this equal to the --max-model-len (as done in the example) or 
slightly higher is standard for maximizing throughput and utilizing the GPU's 
memory bandwidth efficiently for your large context tests. \\
\hline
\textbf{--tensor-parallel-size} & 
\textbf{GPU Parallelism.} Specifies the number of GPUs to split the model across using 
Tensor Parallelism (TP). The model weights and computation are distributed among 
these GPUs. & 
\textbf{Crucial/Necessary (If using a 70B model).} For the Llama-3.1-70B-Instruct (as 
shown in the example), this is essential, as the model weights are too large for 
a single A100 (40GB VRAM). Setting it to 8 implies you are running the experiment 
on an 8-GPU node. Note: If you run your 13B/14B models, you would set this to 1 
on a single A100. \\
\hline
\textbf{--kv-cache-memory-bytes} & 
\textbf{KV Cache Memory Limit.} Explicitly sets the maximum amount of memory (in bytes) 
that vLLM is allowed to use for the KV cache. This is a hard guardrail against 
OOM errors. & 
\textbf{Highly Recommended.} This gives you precise control over memory usage and helps 
prevent the server from crashing. (14,000,000,000 bytes $\approx$ 14 GB). \\
\hline
\end{tabular}
\caption{vLLM server parameters and their necessity for testing}
\label{tab:vllm-parameters}
\end{table}

\subsection{Generated Question Set}
The testing phase utilized custom Python scripts to automate the evaluation of model performance and accuracy across three specific context regimes: 4,096, 10,000, and 15,000 words.

\subsubsection{Core Testing Procedures:}
\begin{itemize}
\item \textbf{Performance and Accuracy Benchmarking:} Each model was presented with a set of 200 general knowledge questions. The scripts captured the execution time (latency) and the accuracy of the responses for each context length.
\item \textbf{Needle-in-a-Haystack (NIAH) Analysis:} Specific facts (the ``needles'') were inserted at the beginning, middle, and end of the distracting context to test the model’s retrieval capabilities and identify potential positional biases.
\item \textbf{Data Archiving:} All results were systematically exported to output files for further statistical analysis.
\end{itemize}

\subsubsection{High-Load Stress Testing}
To observe LLM behavior at peak operational capacity, specifically when GPU utilization exceeded 95\%, two distinct stress-test methodologies were developed:
\begin{itemize}
\item \textbf{Multi-threaded Iteration:} This approach involved increasing the number of concurrent threads until the maximum GPU utilization was reached. This method was found to be most effective for the 4,096-word context regime.
\item \textbf{Multi-loop Execution:} For larger models and higher context lengths (10,000 and 15,000 words), the GPU utilization reached the 95\% threshold almost immediately. To obtain reliable performance averages over a longer duration, we used a 10-loop test rather than simply increasing thread counts.
\end{itemize}

\subsubsection{Experimental Impartiality}
To ensure the test remained unbiased, a set of 200 questions was generated by Google Gemini 2.5 Pro. These questions and their expected answers are cataloged in the Appendix (Table~\ref{tab:questions}). All experimental scripts and the final processed results are available for review in the project’s GitHub repository.

\section{Preliminary Findings for Dense Transformer Models}
The primary focus of the research involved the \textbf{Llama-3.1-70B-Instruct} and \textbf{Qwen1.5-14B-Chat} models. The results across both models highlight the critical importance of context discipline in achieving efficient and reliable LLM performance. The analysis confirms a strong correlation between increased context length and significant performance degradation:

\begin{itemize}
    \item \textbf{Performance Scaling:} Across all models, the inference time exhibited a marked, non-linear increase as the context size scaled from 4,096 words to 10,000 and 15,000 words. This increase was consistently observed in the context-search test, demonstrating that the time required to process the input is dominated by the context size, largely irrespective of the needle’s position.
    \item \textbf{Accuracy Resilience:} The data suggests that while larger context impacts performance universally, the degradation in reasoning accuracy (Context Distraction Test) and retrieval reliability (Needle in a Haystack Test) begins to appear under maximum context loads, underscoring the risk posed by computational pressure on model quality.
\end{itemize}

\subsection{Performance Impact}
\subsubsection{LLM Q\&A}
For this analysis, we selected the second loop (execution) from the baseline test (zero-context) and the 5th execution from the 4,096-, 10,000-, and 15,000-word tests to obtain a like-for-like comparison and avoid initial execution anomalies. For mult-threaded tests, i.e., 4,096-word tests for Qwen, we chose the 2nd execution of the 3rd thread (3, 1) to get to the 5th execution in the following order [(1,0), (2,0), (2,1), (3,0), (3,1), $\ldots$].

Consistent performance degradation across context lengths from the baseline zero-context test to 4,096, 10,000, and 15,000 words is characterized by a marked, non-linear increase in inference time as the system becomes increasingly memory-bandwidth bound.
\begin{figure}[h!t]
  \begin{subfigure}{0.9\linewidth}
  \includegraphics[width=\linewidth]{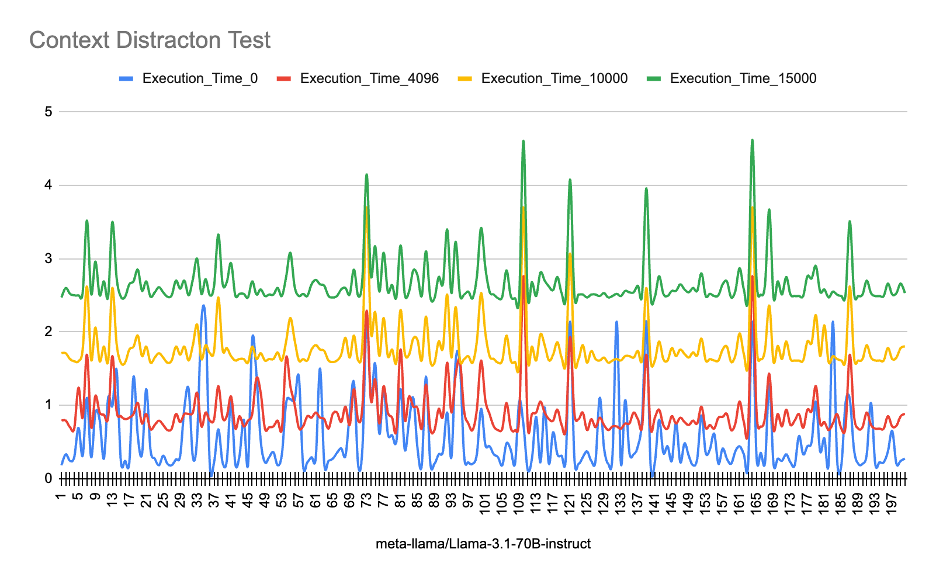}
  \caption{Llama}
  \end{subfigure}\par\medskip
  \begin{subfigure}{0.9\linewidth}
  \includegraphics[width=\linewidth]{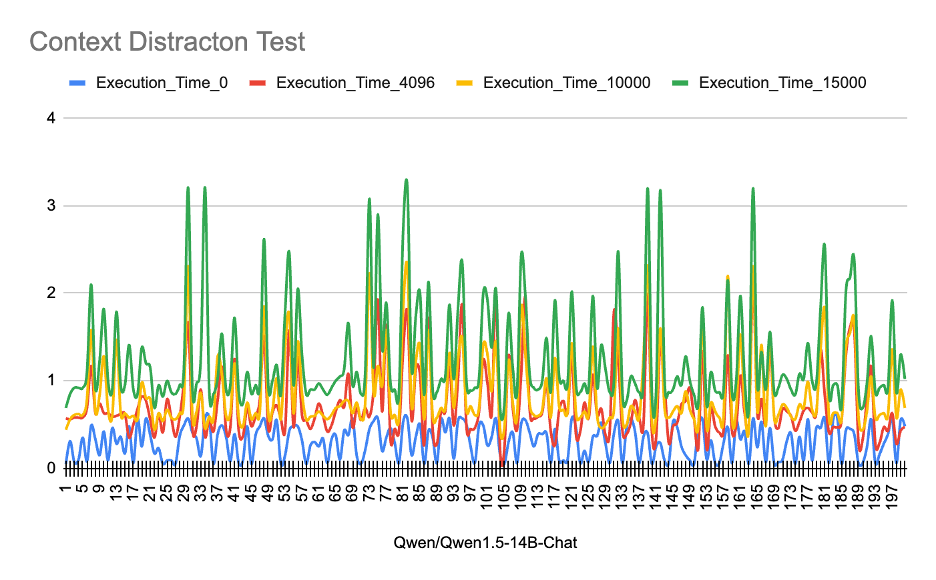}
  \caption{Qwen}
   \end{subfigure}
  \caption{Performance impact}
\end{figure}

To calculate the percentage performance degradation, the following approach is used. Selected five equally distributed outputs from the execution results file, i.e., questions 40, 80, 120, 160, and 200. Calculated the degradation \% with the following formula:
\[
\text{Degradation \%} = \left( \frac{\text{Context-length Execution Time} - \text{Zero-Context Execution Time}}{\text{Zero-Context Execution Time}} \right) \times 100\text{,}
\]
where Context-length execution time refers to the time taken to execute LLM queries for 4,096 words, 10,000 words, and 15,000 words.

Applying the aforementioned formula, we calculated the following performance degradation \% for all evaluated models (with the final row representing the overall average):
\begin{table}[H]
\centering
\begin{tabular}{ccc}
\toprule
\textbf{\% Degradation 4,096} & \textbf{\% Degradation 10,000} & \textbf{\% Degradation 15,000} \\
\midrule
254.16 & 641.66 & 1016.66 \\
32.00 & 216.00 & 394.00 \\
137.93 & 451.72 & 758.62 \\
102.70 & 354.05 & 591.89 \\
225.92 & 566.66 & 837.03 \\
\midrule
\textbf{150.54} & \textbf{446.02} & \textbf{719.64} \\
\bottomrule
\end{tabular}
\caption{Degradation in Llama-3.1-70B-Instruct}
\end{table}

\begin{table}[H]
\centering
\begin{tabular}{ccc}
\toprule
\textbf{\% Degradation 4,096} & \textbf{\% Degradation 10,000} & \textbf{\% Degradation 15,000} \\
\midrule
487.50 & 687.50 & 1087.50 \\
616.66 & 866.66 & 1350.00 \\
530.00 & 550.00 & 870.00 \\
-9.25 & 24.07 & 85.18 \\
-2.08 & 43.75 & 112.50 \\
\midrule
\textbf{324.56} & \textbf{434.39} & \textbf{701.03} \\
\bottomrule
\end{tabular}
\caption{Degradation in Qwen1.5-14B-Chat}
\end{table}

These results indicate that the time required to process the input is dominated by the ``prefill'' stage of the context. As the context length increases, the system becomes memory-bandwidth bound. This suggests that in high-throughput environments, poor ``context discipline''---sending more data than is necessary---acts as a systemic bottleneck that cannot be addressed by increasing computational power alone.

\subsubsection{LLM Context Search}
\begin{figure}[h!tb]
  \begin{subfigure}{0.9\linewidth}
  \includegraphics[width=\linewidth]{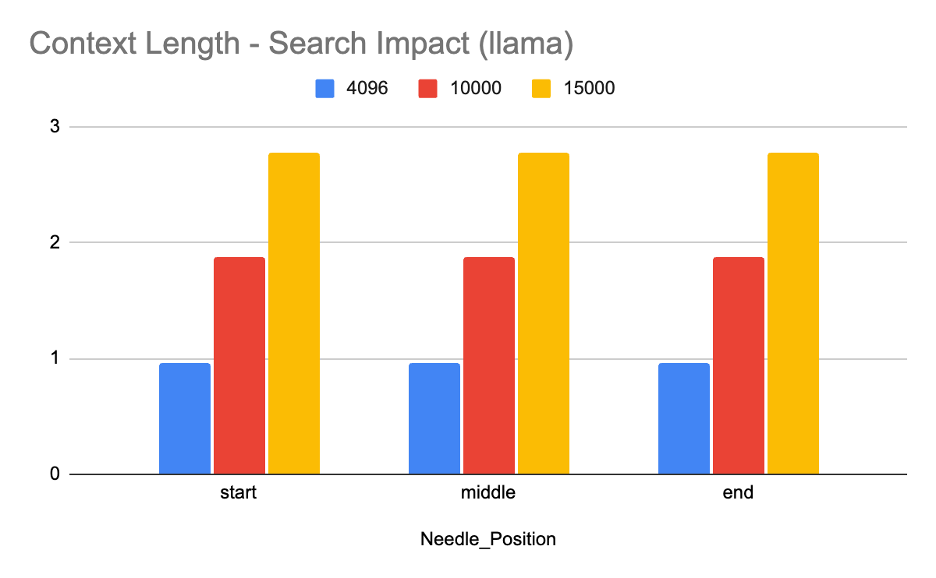}
  \caption{Llama}
  \end{subfigure}\par\medskip
  \begin{subfigure}{0.9\linewidth}
  \includegraphics[width=\linewidth]{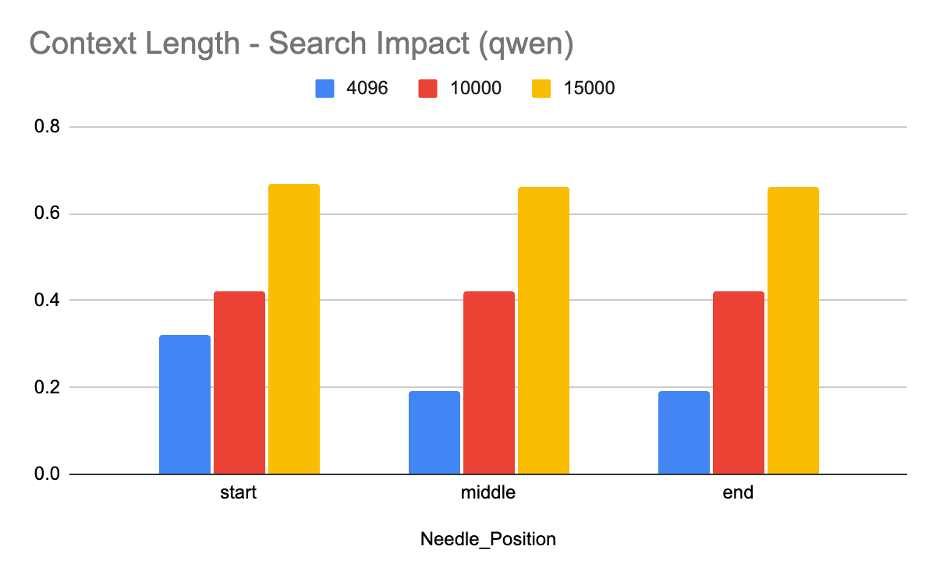}
  \caption{Qwen}
   \end{subfigure}
  \caption{Search impact due to context length}
  \label{fig:search}
\end{figure}

When it comes to the needle in the haystack test, where answers are randomly placed at the start, middle, and end of the context for various context lengths of our research, i.e., 4,096 words, 10,000 words, and 15,000 words, we observe that the models can find the needle irrespective of the context length distraction/noise. However, in terms of performance, a familiar pattern of longer context taking longer to execute is observed, as shown in Fig~\ref{fig:search}.

The observed performance degradation in dense transformer models like Llama-3.1-70B and Qwen1.5-14B follows a linear-quadratic trajectory typical of standard Transformer architectures. As the context expands to 15,000 words, the attention mechanism must perform pairwise comparisons over the entire sequence, significantly increasing the ``prefill'' latency. This results in a ``memory wall'' in which the GPU spends more time moving KV cache data between High Bandwidth Memory (HBM) and SRAM than on actual arithmetic computations, leading to the substantial 719.64\% latency spike observed in the 70B model.

\subsection{Quality Impact of LLM during Q\&A}
When it comes to quality testing, we approached it as two simple tests:
\begin{itemize}
	\item {\textbf{Baseline Quality/accuracy test:} In this test, all models are tested for accuracy against the 200 questions, and the output is compared with the expected output for the questions with zero-context length as baseline. Table~\ref{tab:model-accuracy-1} shows the model accuracy for the baseline test.
	\begin{table}[H]
	\centering
	\begin{tabular}{lcc}
	\toprule
	\textbf{Model} & \textbf{Total Failures} & \textbf{Accuracy \%} \\
	\midrule
	Llama-3.1-70B-Instruct & 3 & 98.5\% \\
	Qwen1.5-14B-Chat & 2 & 99\% \\
	%Mixtral-8x7B-Instruct-v0.1 & 3 & 98.5\% \\
	\bottomrule
	\end{tabular}
	\caption{Model Accuracy and Retrieval Reliability for the Baseline Test}
	\label{tab:model-accuracy-1}
	\end{table}}
	\item {\textbf{Max context-size accuracy test:} Table~\ref{tab:model-accuracy-2} shows the failure rate for 15,000 words context length.
	\begin{table}[H]
	\centering
	\begin{tabular}{lcc}
	\toprule
	\textbf{Model} & \textbf{Total Failures} & \textbf{Accuracy \%} \\
	\midrule
	Llama-3.1-70B-Instruct & 4 & 98\% \\
	Qwen1.5-14B-Chat & 5 & 97.5\% \\
	%Mixtral-8x7B-Instruct-v0.1 & 3 & 98.5\% \\
	\bottomrule
	\end{tabular}
	\caption{Model Accuracy and Error Rates at Maximum Context Capacity (15,000 Words)}
	\label{tab:model-accuracy-2}
	\end{table}}
\end{itemize}

Note that failures are not consistent across different context lengths for the same model. For example, for the Qwen1.5-14B-Chat model, question ID 142 failed the baseline test (zero-context) but passed the other context-length tests at 4,096, 10,000, and 15,000 words.

The accuracy test shows minor degradation between the baseline (zero-context) and the max-context (15,000 words in our testing). This supports the assumption that noisy or invalid context affects LLM accuracy; further testing across different models, more powerful GPUs, and varying context lengths is required to investigate this further. A potential consideration for future research.

The empirical result demonstrates that while increasing context length imposes a severe performance ``tax'' on system resources, the actual reasoning capability of dense transformer models remains remarkably stable. For Llama-3.1-70B, accuracy declined only slightly from the 98.5\% baseline to 98\% at 15,000 words. Similarly, Qwen1.5-14B maintained high performance, shifting only from 99\% to 97.5\%. This resilience suggests a fundamental distinction between the hardware cost of processing context and the model’s ability to maintain ``context discipline''.

\subsection*{Analysis of Resilience and Bottlenecks}
\begin{enumerate}
\item \textbf{IO-Awareness and Memory Walls:} The stability in accuracy suggests that modern attention mechanisms are highly effective at filtering noise, yet the ``memory wall'' increasingly governs them. Dao \textit{et al.}~\cite{dao2022flashattentionfastmemoryefficientexact} argue that the primary bottleneck in long-sequence processing is not computational FLOPs but the reads and writes between slow High Bandwidth Memory (HBM) and fast on-chip SRAM. The latency observed at 15,000 words is a direct consequence of this IO complexity, which remains quadratic ($O(N^2)$) in standard implementations.
\item \textbf{Signal-to-Noise Robustness:} Despite being saturated with 15,000 words of irrelevant ``distractors'', the models demonstrated an ability to rectify or ignore misinformation within the prompt. This aligns with the findings of Aremu \textit{et al.}~\cite{aremu2024reliabilitylargelanguagemodels}, who observed that LLMs can be relied upon to provide correct answers to factual queries even when faced with misinformed or biased prompts. In our tests, the models successfully prioritized the ``needle'' or the core question over the massive volume of irrelevant context.
\item \textbf{Effective Attention vs. Hardware Cost:} The data implies that the ``tax'' for long context is purely operational. While the model’s attention mechanism can successfully ``look away'' from distracting noise to maintain accuracy, the underlying hardware must still load the entire KV cache for every token in the sequence. Consequently, even a ``silent'' or irrelevant token consumes the same memory bandwidth as a critical one, leading to the 719.64\% latency increase observed in the 70B model.
\end{enumerate}

In summary, the models are intellectually robust but operationally expensive under long-context loads. They do not ``lose their way'' as easily as earlier architectures might have. Still, they require significant infrastructure optimization--such as the IO-aware tiling used in FlashAttention~\cite{dao2022flashattentionfastmemoryefficientexact}--to remain viable for real-time applications.

\section{Mixtral Anomaly Analysis}
To investigate the impact of sparse activation, the \textbf{Mixtral-8x7B-Instruct-v0.1} model was subjected to the same benchmarking suite. However, Mistral’s MOE architecture exhibits an anomaly in context: 4,096- and 10,000-word models perform faster than the baseline test (zero-context). An anomaly is observed for the needle placed at the start: the 4,096-word in-context search takes longer than the 10,000- and 15,000-word in-context tests.

The finding that \textit{Mixtral’s response time is slower at 4,096 words than at much longer contexts (10,000 and 15,000 words)}, while Llama-70B and Qwen-14B show expected scaling (latency increasing with context length), strongly suggests an architectural dependency.

\begin{figure}[H]
\centering
\includegraphics[width=0.9\textwidth]{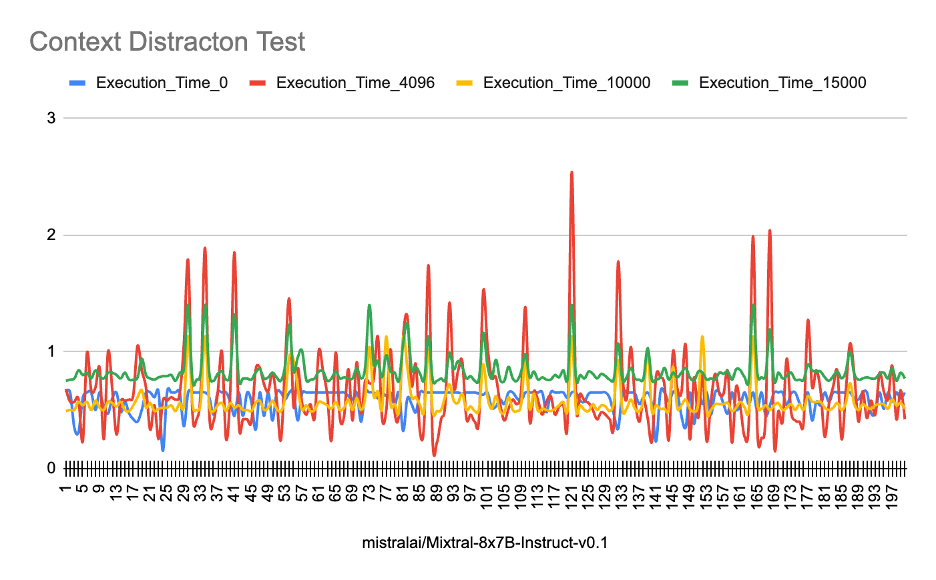}
\caption{Performance impact of Mixtral}
\label{fig:image3}
\end{figure}

Figure~\ref{fig:image3} shows the inconsistent performance degradation across context lengths from the baseline test (zero-context) to 4,096 to 10,000 to 15,000 words, revealing a unique ``hump'' where the 4,096-word regime is unexpectedly slower than longer context lengths due to architectural sparse-routing overhead.

Table~\ref{tab:mixdeg} highlights this performance degradation in percentage.
\begin{table}[H]
\centering
\begin{tabular}{ccc}
\toprule
\textbf{\% Degradation 4,096} & \textbf{\% Degradation 10,000} & \textbf{\% Degradation 15,000} \\
\midrule
7.54 & -1.75 & 43.86 \\
-29.23 & -24.62 & 15.38 \\
-30.00 & -16.67 & 28.33 \\
40.00 & 4.00 & 70.00 \\
-35.38 & -21.54 & 18.46 \\
\midrule
\textbf{-7.41} & \textbf{-12.11} & \textbf{35.21} \\
\bottomrule
\end{tabular}
\caption{Degradation in Mixtral-8x7B-Instruct-v0.1}
\label{tab:mixdeg}
\end{table}

\begin{figure}[H]
\centering
\includegraphics[width=0.9\textwidth]{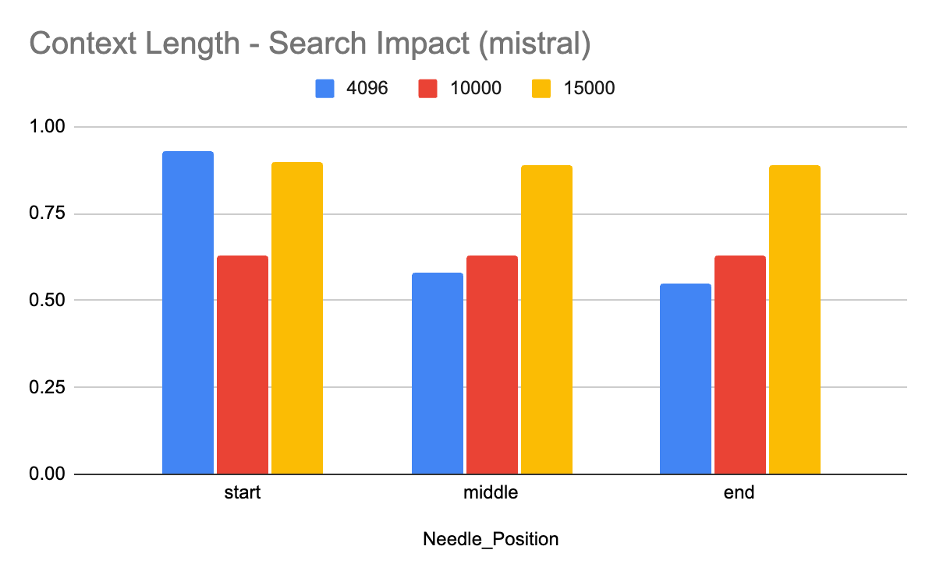}
\caption{Search impact due to context length}
\label{fig:image6}
\end{figure}

Figure~\ref{fig:image6} shows the search impact due to context length for Mixtral.

Table~\ref{tab:model-accuracy-3} shows the accuracy of the model for increased words.
\begin{table}[H]
\centering
\begin{tabular}{lcc}
\toprule
\textbf{Model} & \textbf{Total Failures} & \textbf{Accuracy \%} \\
\midrule
Zero-context & 1 & 99.5\% \\
15,000 words & 3 & 98.5\% \\
\bottomrule
\end{tabular}
\caption{Model performance comparison}
\label{tab:model-accuracy-3}
\end{table}

The observations strongly suggest that this behavior is likely due to the unique \textbf{MoE architecture} of Mixtral, combined with the vLLM serving engine’s handling of the initial ``prefill'' stage.

Here is a detailed explanation of why this anomaly might occur.

\subsection{The MoE Architecture and Sparse Activation}
Mistral’s Mixtral 8x7B model is an MoE, meaning:
\begin{itemize}
	\item \textbf{Total Parameters:} It has about 47 billion parameters in total.
	\item \textbf{Active Parameters:} For any given input words, the router network activates only two of the eight expert layers, resulting in approximately 13 billion active parameters during inference.
\end{itemize}

During inference, the initial prefill stage (which processes the input context) is the most computationally expensive part of the request.

\subsection{The Prefill/Decoding Trade-off}
The two phases of LLM inference are:

\begin{table}[H]
\centering
\begin{tabular}{p{0.15\textwidth}p{0.35\textwidth}p{0.4\textwidth}}
\toprule
\textbf{Phase} & \textbf{Description} & \textbf{Computational Intensity} \\
\midrule
Prefill & 
Processing the input context tokens (e.g., 4,096 words). & 
High (Dominated by matrix multiplication of the full context). \\
\midrule
Decoding & 
Generating the output tokens (e.g., 50 new words). & 
Low (Dominated by memory bandwidth and KV cache management). \\
\bottomrule
\end{tabular}
\caption{Phases of LLM inference and their computational characteristics}
\label{tab:inference-phases}
\end{table}

\subsection{Hypothesis: MoE Prefill Penalty vs. PagedAttention Overhead}
The anomalous slowdown observed in the 4,096-word regime suggests a complex interplay between the Mixture-of-Experts (MoE) architecture and the underlying serving infrastructure. This section explores the two-phase transition where systemic memory bottlenecks eventually overtake architectural routing costs.

\subsubsection{Architectural Routing Penalty and Initial Prefill}
The anomalous latency spike observed at the 4,096-word threshold is likely due to the sparse-activation routing process. During the prefill phase, the model’s gating network must dynamically select and activate two of the eight experts for each incoming token. At moderate context lengths, this routing overhead and the complexity of accessing non-contiguous expert weights in memory impose a significant computational ``tax''. This penalty is evident before the system becomes entirely overwhelmed by raw data volume, creating a localized performance bottleneck unique to the MoE architecture.

\subsubsection{Amortization and Infrastructure Convergence}
As the context scales to 10,000 and 15,000 words, the initial routing penalty becomes secondary to the KV cache’s massive resource requirements. At these extreme scales, the performance bottleneck shifts from model-specific routing to universal infrastructure constraints, specifically the memory bandwidth limits of the vLLM serving engine. The operational cost of loading and computing the KV cache for 15,000 words eventually ``masks'' the MoE-specific overhead. Consequently, the performance of sparse and dense transformer architectures begins to converge, as the same hardware-level ``memory wall'' ultimately restricts both.

\section{Conclusion}
This research provides a definitive quantitative benchmark for the ```context tax'' in modern LLM serving. Our results demonstrate that while high-capacity dense transformer models like Llama-3.1-70B and Qwen1.5-14B exhibit remarkable intellectual resilience---maintaining accuracy levels between 97.5\% and 98.5\% even when saturated with 15,000 words of irrelevant distractors---the operational cost is severe and non-linear. The observed 719.64\% increase in latency for the 70B model at the 15,000-word regime confirms that memory bandwidth and KV cache management are now the primary bottlenecks for long-context applications.

The study also uncovers a critical architectural anomaly in MoE models. The Mixtral-8x7B model exhibited a unique ``hump'' in latency at 4,096 words, suggesting that the computational overhead of sparse routing is most prominent at moderate lengths. However, at extreme scales of 15,000 words, this architectural overhead is entirely masked by the systemic costs of KV cache processing, leading to a performance convergence between sparse and dense transformer architectures.

Ultimately, ``context engineering'', a technical subset of prompt engineering focused on the precise curation and limitation of input data, is no longer merely an optimization preference but a rigorous necessity for maintaining system throughput. As prompt engineering evolves into a broader architectural field, context engineering emerges as the critical link between model performance and hardware constraints. Future advancements must move beyond basic scaling to prioritize IO-aware attention mechanisms and high-efficiency KV cache management, ensuring that model capacity does not outpace practical hardware efficiency.

\bibliographystyle{ieeetr}
\bibliography{paper}

\section*{Appendix}
\subsection*{vLLM Inference Configuration for Qwen and Mixtral Models}

\subsubsection*{Qwen1.5-14B-Chat}
\begin{Verbatim}[fontsize=\footnotesize]
python3 -m vllm.entrypoints.openai.api_server \
    --model Qwen/Qwen1.5-14B-Chat \
    --max-model-len 32768 \
    --host 0.0.0.0 \
    --port 8000 \
    --no-enable-prefix-caching \
    --max-num-batched-tokens 32768 \
    --tensor-parallel-size 8 \
    --kv-cache-memory-bytes 14000000000
\end{Verbatim}

\subsubsection*{Mixtral-8x7B-Instruct-v0.1}
\begin{Verbatim}[fontsize=\footnotesize]
python3 -m vllm.entrypoints.openai.api_server \
    --model mistralai/Mixtral-8x7B-Instruct-v0.1 \
    --max-model-len 32768 \
    --host 0.0.0.0 \
    --port 8000 \
    --no-enable-prefix-caching \
    --max-num-batched-tokens 32768 \
    --tensor-parallel-size 8 \
    --kv-cache-memory-bytes 14000000000
\end{Verbatim} 

\subsection*{Gemini Pro 2.5 Generated Questions for Testing}

\begin{longtable}{cp{0.55\textwidth}p{0.25\textwidth}}
\toprule
\footnotesize
\textbf{ID} & \textbf{Question} & \textbf{Expected Answer} \\ \hline
\endfirsthead

\toprule
\textbf{ID} & \textbf{Question} & \textbf{Expected Answer} \\ \hline
\endhead

\hline
\endfoot

\endlastfoot

1 & What is the capital of France? & paris \\ \hline
2 & Who wrote `Hamlet'? & shakespeare \\ \hline
3 & What is the chemical symbol for water? & h2o \\ \hline
4 & In which year did the Titanic sink? & 1912 \\ \hline
5 & What planet is known as the Red Planet? & mars \\ \hline
6 & Who painted the Mona Lisa? & vinci \\ \hline
7 & What is the tallest mountain in the world? & everest \\ \hline
8 & What is the main ingredient in guacamole? & avocado \\ \hline
9 & How many continents are there? & seven \\ \hline
10 & Who was the first person to walk on the moon? & armstrong \\ \hline
11 & What is the currency of Japan? & yen \\ \hline
12 & What is the hardest natural substance on Earth? & diamond \\ \hline
13 & Which ocean is the largest? & pacific \\ \hline
14 & Who invented the telephone? & bell \\ \hline
15 & What is the square root of 64? & 8 \\ \hline
16 & Which country is famous for its pyramids? & egypt \\ \hline
17 & What is the primary language spoken in Brazil? & portuguese \\ \hline
18 & Who discovered penicillin? & fleming \\ \hline
19 & What is the boiling point of water at sea level? & 100 \\ \hline
20 & Which artist cut off his own ear? & van gogh \\ \hline
21 & What is the largest animal in the world? & blue whale \\ \hline
22 & In what country would you find the Eiffel Tower? & france \\ \hline
23 & What is the name of the galaxy we live in? & milky way \\ \hline
24 & How many sides does a triangle have? & three \\ \hline
25 & Who is the author of the Harry Potter series? & rowling \\ \hline
26 & What is the chemical symbol for gold? & au \\ \hline
27 & What is the capital of Italy? & rome \\ \hline
28 & Who was the first President of the United States? & washington \\ \hline
29 & What gas do plants absorb from the atmosphere? & carbon dioxide \\ \hline
30 & In which decade was the Internet invented? & 1960s \\ \hline
31 & What is the largest planet in our solar system? & jupiter \\ \hline
32 & Who wrote `To Kill a Mockingbird'? & lee \\ \hline
33 & What is the currency of China? & yuan \\ \hline
34 & What is the process by which plants make their food? & photosynthesis \\ \hline
35 & Which animal is the largest primate? & gorilla \\ \hline
36 & What is the capital of Spain? & madrid \\ \hline
37 & What color is a ruby? & red \\ \hline
38 & What is the longest river in the world? & nile \\ \hline
39 & Who developed the theory of relativity? & einstein \\ \hline
40 & How many states are there in the United States? & 50 \\ \hline
41 & What is the highest-grossing film of all time (original release)? & avatar \\ \hline
42 & What is the chemical symbol for oxygen? & o \\ \hline
43 & Which element has the atomic number 1? & hydrogen \\ \hline
44 & What is the name of the main villain in the `Lord of the Rings' trilogy? & sauron \\ \hline
45 & What is the capital of Germany? & berlin \\ \hline
46 & What is the largest organ in the human body? & skin \\ \hline
47 & Who painted the ceiling of the Sistine Chapel? & michelangelo \\ \hline
48 & What is the primary gas in Earth's atmosphere? & nitrogen \\ \hline
49 & What year did World War I begin? & 1914 \\ \hline
50 & Which instrument is used to measure earthquakes? & seismograph \\ \hline
51 & What is the name of the fictional detective who lives at 221B Baker Street? & holmes \\ \hline
52 & What is the capital of Russia? & moscow \\ \hline
53 & What is the chemical symbol for salt? & nacl \\ \hline
54 & Who was the first woman to win a Nobel Prize? & curie \\ \hline
55 & What is the largest desert in the world? & sahara \\ \hline
56 & What metal is the best conductor of electricity? & silver \\ \hline
57 & What is the name of the famous clock tower in London? & big ben \\ \hline
58 & What is the capital of Canada? & ottawa \\ \hline
59 & Which planet is closest to the sun? & mercury \\ \hline
60 & Who wrote `1984'? & orwell \\ \hline
61 & What is the unit of electric current? & ampere \\ \hline
62 & What is the world's most populous country? & china \\ \hline
63 & What is the chemical symbol for iron? & fe \\ \hline
64 & Who was the Greek god of the sea? & poseidon \\ \hline
65 & How many bones are in the adult human body? & 206 \\ \hline
66 & What city hosted the 2000 Summer Olympics? & sydney \\ \hline
67 & What famous German composer wrote the `Ninth Symphony'? & beethoven \\ \hline
68 & What is the main chemical compound in natural gas? & methane \\ \hline
69 & Which war ended in 1945? & world war ii \\ \hline
70 & What is the main character's name in `The Great Gatsby'? & gatsby \\ \hline
71 & What is the capital of Australia? & canberra \\ \hline
72 & What is the boiling point of the Celsius scale? & 100 \\ \hline
73 & Who invented the light bulb? & edison \\ \hline
74 & What type of star is the sun? & g-type main-sequence \\ \hline
75 & What is the formula for calculating force? & f=ma \\ \hline
76 & What is the name of the first satellite launched into space? & sputnik \\ \hline
77 & What is the smallest country in the world? & vatican city \\ \hline
78 & Who painted `The Starry Night'? & van gogh \\ \hline
79 & What is the largest country by land area? & russia \\ \hline
80 & What is the chemical symbol for silver? & ag \\ \hline
81 & Who was the Queen of England for the longest time? & elizabeth ii \\ \hline
82 & What is the process of a liquid turning into a gas? & evaporation \\ \hline
83 & What is the name of the currency used in the United Kingdom? & pound sterling \\ \hline
84 & What is the largest moon in our solar system? & ganymede \\ \hline
85 & Who composed `The Marriage of Figaro'? & mozart \\ \hline
86 & What is the capital of China? & beijing \\ \hline
87 & What is the primary function of red blood cells? & oxygen transport \\ \hline
88 & What is the study of living organisms called? & biology \\ \hline
89 & What is the chemical symbol for carbon? & c \\ \hline
90 & Which mountain range runs along the western coast of South America? & andes \\ \hline
91 & Who wrote `The Odyssey'? & homer \\ \hline
92 & What is the fastest animal on Earth? & cheetah \\ \hline
93 & What is the freezing point of water in Fahrenheit? & 32 \\ \hline
94 & What major historical event occurred in 1776? & american independence \\ \hline
95 & What material is a pencil's lead made of? & graphite \\ \hline
96 & Who discovered the laws of motion and universal gravitation? & newton \\ \hline
97 & What is the name of the strait that separates Europe and Africa? & gibraltar \\ \hline
98 & What is the chemical symbol for potassium? & k \\ \hline
99 & Who was the leader of the Soviet Union during World War II? & stalin \\ \hline
100 & What is the deepest point in the Earth's oceans? & mariana trench \\ \hline
101 & What is the largest body of water in the world that is not an ocean? & caspian sea \\ \hline
102 & Who composed `Four Seasons'? & vivaldi \\ \hline
103 & What is the process of turning a gas directly into a solid called? & deposition \\ \hline
104 & Which country is home to the Kangaroo? & australia \\ \hline
105 & What is the capital of Japan? & tokyo \\ \hline
106 & What is the hardest material on the Mohs scale? & diamond \\ \hline
107 & Who was the Ancient Greek mathematician known as the `Father of Geometry'? & euclid \\ \hline
108 & What is the chemical symbol for copper? & cu \\ \hline
109 & What is the main component of Earth's core? & iron and nickel \\ \hline
110 & What famous phrase did Julius Caesar reportedly say when crossing the Rubicon? & alea iacta est \\ \hline
111 & What is the name of the ship that brought the Pilgrims to America? & mayflower \\ \hline
112 & What is the unit of frequency? & hertz \\ \hline
113 & What planet is known for its rings? & saturn \\ \hline
114 & Who invented the printing press? & gutenberg \\ \hline
115 & What is the most abundant element in the universe? & hydrogen \\ \hline
116 & What is the capital of Mexico? & mexico city \\ \hline
117 & Which classical composer was deaf? & beethoven \\ \hline
118 & What is the name of the currency used in Russia? & ruble \\ \hline
119 & What is the chemical formula for sulfuric acid? & h2so4 \\ \hline
120 & In which city is the Colosseum located? & rome \\ \hline
121 & What are the two major components of a simple battery? & anode and cathode \\ \hline
122 & What city is known as `The Big Apple'? & new york \\ \hline
123 & What is the capital of Brazil? & brasilia \\ \hline
124 & Who wrote `Pride and Prejudice'? & austen \\ \hline
125 & What is the common name for sodium chloride? & salt \\ \hline
126 & How many days are in a leap year? & 366 \\ \hline
127 & Who was the Greek goddess of wisdom and war? & athena \\ \hline
128 & What is the largest lake in Africa? & victoria \\ \hline
129 & Who was the German philosopher known for `Thus Spoke Zarathustra'? & nietzsche \\ \hline
130 & What is the chemical symbol for calcium? & ca \\ \hline
131 & What is the process of cell division in body cells called? & mitosis \\ \hline
132 & What is the main type of rock found at the bottom of the ocean? & sedimentary \\ \hline
133 & What is the capital of South Korea? & seoul \\ \hline
134 & What famous landmark was built by the Emperor Shah Jahan? & taj mahal \\ \hline
135 & What is the main pigment responsible for the green color in plants? & chlorophyll \\ \hline
136 & Who composed the opera `Carmen'? & bizet \\ \hline
137 & What is the chemical symbol for tin? & sn \\ \hline
138 & What is the name of the supercontinent that existed millions of years ago? & pangea \\ \hline
139 & Who invented the steam engine? & watt \\ \hline
140 & What is the capital of Egypt? & cairo \\ \hline
141 & What is the primary language spoken in Mexico? & spanish \\ \hline
142 & Who was the first female Prime Minister of the United Kingdom? & thatcher \\ \hline
143 & What is the common name for the Aurora Borealis? & northern lights \\ \hline
144 & What is the chemical symbol for lead? & pb \\ \hline
145 & What is the unit of power? & watt \\ \hline
146 & Which artist is famous for cutting his ear off? & van gogh \\ \hline
147 & What is the largest country in South America? & brazil \\ \hline
148 & What is the chemical formula for table sugar? & c12h22o11 \\ \hline
149 & Who was the author of `Moby Dick'? & melville \\ \hline
150 & What is the capital of Ireland? & dublin \\ \hline
151 & What type of rock is formed from cooling magma? & igneous \\ \hline
152 & Who was the first Roman Emperor? & augustus \\ \hline
153 & What is the common name for the disease poliomyelitis? & polio \\ \hline
154 & What is the name of the imaginary line that divides the Earth into North and South? & equator \\ \hline
155 & What is the chemical symbol for mercury? & hg \\ \hline
156 & What is the capital of New Zealand? & wellington \\ \hline
157 & Who painted `Guernica'? & picasso \\ \hline
158 & What is the smallest ocean in the world? & arctic \\ \hline
159 & What is the chemical symbol for neon? & ne \\ \hline
160 & What famous document was signed in 1776? & declaration of independence \\ \hline
161 & What are the three primary colors of light? & red, green, blue \\ \hline
162 & Who is the author of `The Canterbury Tales'? & chaucer \\ \hline
163 & What is the capital of Greece? & athens \\ \hline
164 & What is the process of generating energy from the sun? & solar power \\ \hline
165 & What is the largest internal organ in the human body? & liver \\ \hline
166 & What is the name of the Russian space agency? & roscosmos \\ \hline
167 & What is the main language spoken in Portugal? & portuguese \\ \hline
168 & Who was the longest-reigning British monarch? & elizabeth ii \\ \hline
169 & What is the chemical symbol for sodium? & na \\ \hline
170 & What is the unit of resistance? & ohm \\ \hline
171 & In what country would you find Mount Kilimanjaro? & tanzania \\ \hline
172 & Who wrote `The Communist Manifesto'? & marx \\ \hline
173 & What is the process of burning fuel called? & combustion \\ \hline
174 & What is the capital of Sweden? & stockholm \\ \hline
175 & What is the smallest planet in the solar system? & mercury \\ \hline
176 & What element is denoted by the chemical symbol `Sn'? & tin \\ \hline
177 & Who designed the Statue of Liberty? & bartholdi \\ \hline
178 & What is the currency of India? & rupee \\ \hline
179 & What is the most widely spoken language in the world? & mandarin chinese \\ \hline
180 & What is the name of the imaginary line that passes through Greenwich? & prime meridian \\ \hline
181 & What famous ship was sunk by a German U-boat in 1915? & lusitania \\ \hline
182 & What is the chemical symbol for chlorine? & cl \\ \hline
183 & Who invented the World Wide Web? & tim berners-lee \\ \hline
184 & What is the outer layer of the Earth called? & crust \\ \hline
185 & What is the capital of Thailand? & bangkok \\ \hline
186 & What type of energy is stored in a compressed spring? & potential \\ \hline
187 & What is the deepest lake in the world? & baikal \\ \hline
188 & Who painted `The Scream'? & munch \\ \hline
189 & What is the chemical symbol for zinc? & zn \\ \hline
190 & What is the capital of Saudi Arabia? & riyadh \\ \hline
191 & Who composed `Clair de Lune'? & debussy \\ \hline
192 & What is the name of the process that powers the sun? & nuclear fusion \\ \hline
193 & What is the capital of Nigeria? & abuja \\ \hline
194 & What is the chemical formula for methane? & ch4 \\ \hline
195 & What year did the Berlin Wall fall? & 1989 \\ \hline
196 & Who wrote `War and Peace'? & tolstoy \\ \hline
197 & What is the main gas in Venus' atmosphere? & carbon dioxide \\ \hline
198 & What is the capital of Argentina? & buenos aires \\ \hline
199 & What is the pH level of neutral water? & 7 \\ \hline
200 & Who developed the first successful printing press? & gutenberg \\ \bottomrule
\caption{Questions for testing}
\label{tab:questions}
\end{longtable}

\end{document}